\documentclass[a4paper, 10pt, conference]{ieeeconf}      

\IEEEoverridecommandlockouts                              

\usepackage{graphics} 

\title{\LARGE \bf
Artifacts Mapping: Multi-Modal Semantic Mapping for Object Detection and 3D Localization
}

\author{
Federico Rollo\textsuperscript{$\dagger$,$\ddagger$,$\S$},
Gennaro Raiola\textsuperscript{$\dagger$,$\ddagger$},
Andrea Zunino\textsuperscript{$\dagger$,$\ddagger$},
Nikolaos Tsagarakis\textsuperscript{$\ddagger$},
Arash Ajoudani\textsuperscript{$\ddagger$} 
\thanks{\textsuperscript{$\dagger$}Intelligent and Autonomous Systems, Leonardo Labs, Genoa, Italy}
\thanks{\textsuperscript{$\ddagger$}HHCM \& HRII, Istituto Italiano di Tecnologia, Genoa, Italy}%
\thanks{\textsuperscript{$\S$}Industrial Innovation, DISI, Università di Trento, Trento, Italy}%
\thanks{Authors' e-mail: \textit{\{name.surname\}.ext@leonardo.com}}
}

\usepackage{graphicx}
\usepackage{amsmath}
\usepackage{amssymb}
\usepackage{url}

\makeatletter
\let\NAT@parse\undefined
\makeatother
\usepackage[colorlinks=true,allcolors=red]{hyperref}

\usepackage{xcolor}
\usepackage{soul}

\begin{document}


\maketitle


\begin{abstract} \label{sect:abstract}
    Geometric navigation is nowadays a well-established field of robotics and the research focus is shifting towards higher-level scene understanding, such as Semantic Mapping. 
When a robot needs to interact with its environment, it must be able to comprehend the contextual information of its surroundings. 
This work focuses on classifying and localising objects within a map, which is under construction (SLAM) or already built. 
To further explore this direction, we propose a framework that can autonomously detect and localize predefined objects in a known environment using a multi-modal sensor fusion approach (combining RGB and depth data from an RGB-D camera and a lidar). 
The framework consists of three key elements: understanding the environment through RGB data, estimating depth through multi-modal sensor fusion, and managing artifacts (\textit{i.e.}, filtering and stabilizing measurements). 
The experiments show that the proposed framework can accurately detect $98\%$ of the objects in the real sample environment, without post-processing, while $85\%$ and $80\%$ of the objects were mapped using the single RGBD camera or RGB + lidar setup respectively. The comparison with single-sensor (camera or lidar) experiments is performed to show that sensor fusion allows the robot to accurately detect near and far obstacles, which would have been noisy or imprecise in a purely visual or laser-based approach.

\end{abstract}

\section{Introduction} \label{sect:intro}


To boost navigation autonomy and contextual awareness of mobile robots in unstructured environments, geometric information collected from the surroundings and the associated semantic data play key roles. The latter, in particular, includes qualitative environment information that can contribute to improving the robot's autonomy for navigation, task planning and manipulation, and simplifying human-robot interaction (HRI). This problem is tackled in the \textit{Semantic Mapping} field, which aims to organize objects into classes and compute their pose and shape in a specific fixed reference frame. In this way, the environmental geometric information is supported by high-level features which increase the robot's awareness of the environment. 
In our specific case, we deal with the object detection and localization problem, which nowadays is widely investigated. For instance, in the last Darpa Subterranean Challenge\footnote{Darpa Subterranean Challenge: \url{https://www.subtchallenge.com/}}, the main objectives were multi-robot exploration and object mapping in unknown environments, and the overall score was calculated based on the number of correctly detected and localized objects on the map. 

\begin{figure}[t]
    \centering
    \includegraphics[width=\linewidth]{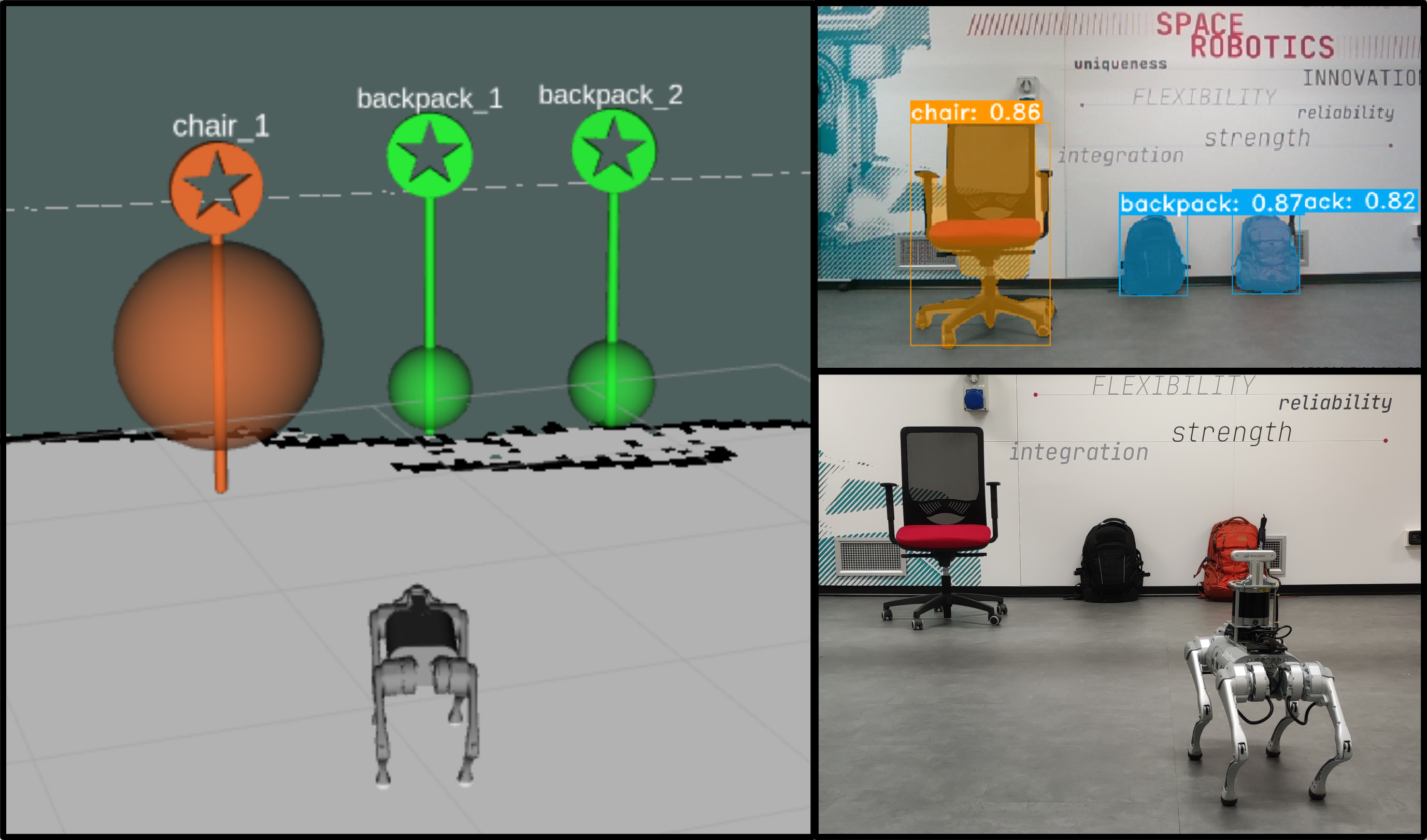}
    \caption{An example of the framework during an experiment. On the left, is the visual application where objects are shown with a landmark and a spherical region of interest for the location in the Rviz visualization tool. On the top right, is the instance segmentation inference of the image taken from the robot camera while on the bottom right is the external representation of the experimental scene.}
    \label{fig:artifact_example}
\end{figure}


Different works were proposed to cope with the semantic mapping problem. Most recent results in robotics are facing the problem of using only RGB data and some interactive structures to be compliant with dynamic environments \cite{hughes2022hydra} while others rely on RGB-D data exploiting older algorithmic strategies (\textit{e.g.} PnP algorithm) \cite{hau2022object}. In autonomous driving, the RGB camera and lidar sensor fusion for semantic understanding is a currently tackled problem~\cite{liang2019multi}. For a broader evaluation of the literature review see Sect.~\ref{sec:works}.

Independently of the approaches used in robotics literature, the first thing which stands out is that most of them rely only on camera sensors. Cameras can give lots of dense information to the user especially if paired with depth data. However, their accurate depth range is within a few meters, leading to heavy depth measurement errors as the distances increase, especially if the robot is moving. This is particularly true for outdoor and vast indoor environments (\textit{e.g.}, warehouses), where depth cameras are limiting and object semantic mapping remains a major challenge for far distances. In these cases, lidar sensors are an essential camera partner, allowing to have precise depth measurements for a wider distance range. Rather, in autonomous driving, the lidar and the RGB camera are nowadays commonly used but depth cameras are not considered due to their low resolution in the wide outdoor areas commonly faced in driving scenarios.

\begin{figure}[t]
    \centering
    \includegraphics[width=\linewidth]{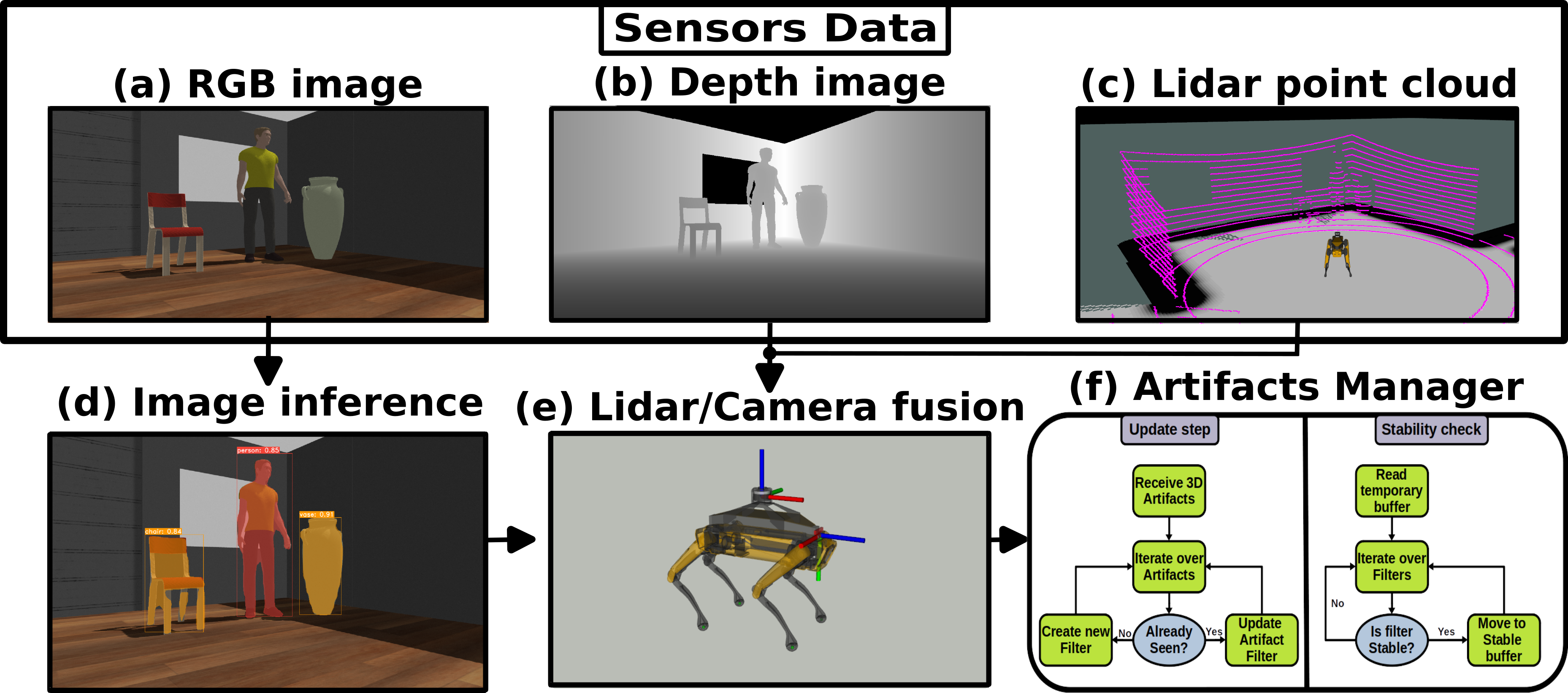}
    \caption{This figure represents the whole Artifacts Mapping pipeline. The top block groups the sensors' data readings: (a) camera RGB image, (b) camera depth image and (c) lidar point cloud. At the bottom, there are (d) the RGB image inference performed with a Deep Neural Network for instance segmentation, (e) the multi-modal sensor fusion for detection and localization which uses as input the camera depth, the lidar point cloud and the Neural Network inference, and (f) a representation of the artifacts manager state-machines used to handle the sensor fusion detections and stabilize them.}
    \label{fig:pipeline}
\end{figure}

Another aspect not considered in most of the robotics examined works is that they do not account for limited resources applications which should run on embedded devices (\textit{e.g.} Nvidia Jetson Nano\footnote{Nvidia Jetson Nano:~\url{https://developer.nvidia.com/embedded/jetson-nano-developer-kit}}). Furthermore, Semantic Mapping is often used in the context of grasping or augmented reality scenarios while this work proposes an application for detecting and localizing objects (a.k.a artifacts) for high-level navigation tasks.


In our work, we aim to merge robotics and autonomous driving applications' strengths and present a modular architecture for semantic mapping\footnote{Artifacts Mapping Youtube videos: \url{https://www.youtube.com/playlist?list=PLdibjJfM06zugiWd-yUcdGH-SRWKTA3nQ}}. We provide a multi-modal (camera-lidar) online semantic mapping framework which can fuse sensor information in real-time depending on the object distance and sensor's accuracies. We use image semantic information to enrich objects' filtered and stabilised positions to have precise object localization. The artifacts' shapes are simplified as spheres but they will be improved in future development. Our work relies on external geometric-based navigation frameworks such as SLAM algorithms or other localization algorithms (\textit{e.g.}, AMCL~\cite{xiaoyu2018adaptive}).


The proposed application demonstrates good accuracy for both near and far objects thanks to the camera-lidar depth fusion which, as far as the authors know, was not examined in other robotic or autonomous driving semantic mapping works. The application operates online also on low resources embedded systems (see Sect.~\ref{subec:realexp}) which strengthens the contributions of this paper. 
Moreover, we developed a Rviz\footnote{rviz:~\url{http://wiki.ros.org/rviz}} application which improves the user experience (UX) for visualization and interaction with the objects and the robot (see Fig.~\ref{fig:artifact_example}).
The authors provide the source code at \url{https://github.com/FedericoRollo/artifacts_mapping} as an added contribution, for running the artifacts mapping applications in simulation or on a robot (see Sect~\ref{subsect:perception}-\ref{subsect:manager} and Sect.~\ref{subsect:application}).

The paper is divided as follows. In Sect.~\ref{sec:works} a literature review of some recent works in semantic mapping is presented. The framework developed for this work is explained in Sect.~\ref{sect:artifacts}. We can distinguish the framework pipeline as two perception modules and a manager one. The first perception module performs 2D object detection while the second aims to estimate 3D artifacts position by fusing camera and lidar depth information (see Sect.~\ref{subsect:perception}). The last module is needed to stabilize the perception estimations and to filter out noisy outliers (see Sect.~\ref{subsect:manager}). An application of the presented framework (see Sect.~\ref{subsect:application}) is proposed based on two steps: \textit{(i)} the robot can autonomously classify and localize objects on a map and save them in a specified format, \textit{(ii)} the robot can load the artifacts as way-points on the map and the user can interactively select them to command the robot moving in that place to successively accomplish various kind of tasks such as manipulation, grasping, inspections or others. In Sect.~\ref{sect:experiments} the experiments to validate the framework are evaluated and discussed, and in Sect.~\ref{sect:conclusion} the conclusion and some future improvements are provided.

\section{Related works} \label{sec:works}


In literature, the semantic mapping problem was addressed using several approaches both in robotics and autonomous driving fields. Different surveys were presented that analysed this topic from various points of view. In \cite{achour2022collaborative} the authors explored the semantic mapping application in a human-robot collaboration scenario in an indoor environment while in \cite{xia2020survey} the semantic SLAM problem is presented in a general fashion analysing the works also in terms of perception, robustness, and accuracy. In \cite{kostavelis2015semantic}, the less recent semantic mapping works are reviewed (\textit{i.e.}, before 2014). This survey is a good reference to analyse the first development for the semantic mapping problem which yielded the more recent applications. 

Among the modern semantic mapping approaches presented in robotics literature in the last decade, some first successful examples are \cite{civera2011towards} and \cite{salas2013slam++}. In \cite{civera2011towards} the authors presented a monocular SLAM system that uses a SURF~\cite{bay2008speeded} feature extractor to check correspondencies and reconstruct the object's geometry. Instead, the authors in \cite{salas2013slam++} showed an object-oriented 3D SLAM based on an ICP~\cite{besl1992method} object pose refinement and demonstrated that the introduction of semantic objects in the SLAM loop improves performances. the authors in \cite{pillai2015monocular} developed a monocular SLAM-aware object recognition system based on multi-view object proposals and efficient feature encoding methods giving as output a semi-dense semantic map. In \cite{tateno20162} the authors proposed a framework which directly manages 3D objects. They use a Kinect\footnote{Microsoft Kinect camera~\url{https://en.wikipedia.org/wiki/Kinect}} camera to reconstruct the 3D environment from different points of view and classify them while estimating their pose. In \cite{xiang2017rnn} the Data Associated Recurrent Neural Networks (DA-RNN) is introduced, which is an RNN for semantic labelling of RGB-D videos. The network output is fused with the KinectFusion algorithm~\cite{newcombe2011kinectfusion} to merge semantic and geometric data. In \cite{mccormac2017semanticfusion} a Convolutional Neural Network (CNN) is used along with the ElasticFusion SLAM algorithm~\cite{whelan2015elasticfusion} to provide long-term dense correspondences between RGB-D video frames even in loopy trajectories. The authors in \cite{sunderhauf2017meaningful} leveraged ORB-SLAM2~\cite{mur2017orb} to reconstruct the geometric environment while using Single-Shot multi-box Detector (SSD)~\cite{liu2016ssd} along with an unsupervised 3D segmentation algorithm to place objects in the environment. 

Moving towards more recent works, in \cite{zeng2018semantic} is presented the Contextual Temporal Mapping (CT-Map). They modelled the semantic inference as a Conditional Random Field (CRF) to account for contextual relations between objects and the temporal consistency of their pose. MaskFusion~\cite{runz2018maskfusion} is a real-time object-aware semantic and dynamic RGB-D SLAM algorithm. The greatest difference with respect to its predecessors is that it can cope with dynamic objects by continuously labelling them. Fusion++~\cite{mccormac2018fusion} performs an object-level SLAM based on a 3D graph map of arbitrary reconstructed objects. They used RGB-D cameras, Mask-RCNN~\cite{he2017mask} instance segmentation and the Truncated Signed Distance Function (TSDF) to perform the semantic reconstruction. 
In \cite{grinvald2019volumetric} is presented an approach that incrementally builds a volumetric object-centric map with an RGB-D camera. They used an unsupervised geometric approach with instance-aware semantic predictions to detect previously unseen objects. They then associated the 3D shape locations with their classes if available and integrate them into the map. This approach has limited time performances to be used on a mobile robot because it runs at 1 HZ so it could be impractical in real-time. Conversely, in \cite{pham2019real} the authors obtained a real-time dense reconstruction and semantic segmentation of 3D indoor scenes. They used an efficient super-voxel clustering method and conditional random fields (CRF) with higher order constraints from structural and object cues, enabling progressive dense semantic segmentation without any precomputation. The CRF infer optimal segmentation labels from the prediction of a deep neural network and runs in parallel with a real-time 3D reconstructor which utilizes RGB-D images as input.
In \cite{rosinol2020kimera} an open-source C++ library for metric-semantic visual-inertial SLAM in real-time is presented. They provide a modular code composed of a visual-inertial odometry (VIO) module, a pose graph optimizer, a 3D mesh-building module, and a dense 3D metric-semantic reconstruction module.
The authors in \cite{bultmann2021real}, used a UAV equipped with a lidar, an RGB camera and a thermal camera to augment 3D point clouds and image segmentation masks while also generating an allocentric map.

One of the last available works which focus on this topic is \cite{hughes2022hydra} which presented a semantic mapping framework which uses only RGB data. They did not accomplish only object mapping but they provided a framework that can also distinguish different rooms and buildings. They exploited the 3D dynamic scene graphs~\cite{rosinol20203d} to abstract the different layers of inference (\textit{i.e.} object, room and building), to solve problems such as loop closure detection and to cope with the mapping problem. 
Instead, the authors of \cite{hau2022object} used RGB-D cameras to reconstruct an allocentric semantic map. They used a keypoint-based approach for pose estimation using a CNN keypoint extractor trained on synthetic data. Object poses were recovered from keypoint detections in each camera viewpoint with a variant of the PnP algorithm. The outputs obtained from the multi-camera system were then fused using weighted interpolation.

In autonomous driving, the multi-sensor fusion problem for 3D object detection is faced in \cite{liang2019multi} which uses lidar and RGB camera sensors to estimate the objects positions in the environment through ground estimation and depth completion. They use an end-to-end approach to train their multi-task network.
The authors in \cite{chen2019suma++} build a semantic map with a laser-based semantic segmentation of the point cloud not requiring any camera data.
In \cite{li2020building}, the authors provided a lidar-based SLAM for the geometric mapping and then use a CRF to fuse and optimize the camera semantic labels to obtain the semantic map. Instead, in \cite{berrio2021camera}, the camera and lidar data are used to build a probabilistic semantic octree map considering all the uncertainties of the sensors involved in the process.
The authors in \cite{cheng2022vision} presented one of the latest works in autonomous driving semantic mapping. They use an RGB camera and a lidar to perform semantic segmentation, direct sparse visual odometry and global optimization to include GNSS data in the mapping process. 

Our review of the state-of-the-art indicated that most of the works on robotics platforms rely only on camera measurements and the experiments are limited to small indoor environments. Instead, in the autonomous driving scenario the camera-lidar fusion is already used for semantic tasks but they rarely use depth cameras, their lidars are generally more powerful (\textit{i.e.}, they have 128-row lidars compared to the 16 ones commonly used in robotics) and they test the application in driving outdoor scenarios which offer different challenges with respect to robotic indoor once. Hence, with our work, we aim to stress the fact that RGB-D cameras and lidars are complementary sensors also in robotic semantic applications. For the semantic mapping application, we stated that with both sensors we can correctly localize objects at different distance ranges, improving detection accuracy.

\section{Artifact Mapping Framework} \label{sect:artifacts}

In this section, the whole framework is presented as a conjunction of two blocks: Sect.~\ref{subsect:perception} for object perception and Sect.~\ref{subsect:manager} for object managing. In Sect.~\ref{subsect:application} the provided UI application is illustrated.  

\subsection{Artifacts detection and position estimation} \label{subsect:perception}
The perception part can be conceptually divided into two components: \textit{(i)} 2D object segmentation, \textit{(ii)} 3D object position estimation using camera-lidar filtering.

\subsubsection{2D object segmentation}
In this phase, a deep neural network~\cite{bolya2020yolact++} is used to infer from RGB images (see Fig.~\ref{fig:pipeline}a) some predefined objects' classes and their masks. During the navigation, the robot takes pictures of the environment using the camera mounted on it. The pictures are passed into an instance segmentation deep neural network which outputs the classification labels and masks (\textit{i.e.}, a binary image having 1 where the object is found and 0 elsewhere) for each object recognized on the image (see Fig.~\ref{fig:pipeline}d).
The outputs are grouped and passed to the next module which will convert 2D data into 3D ones. 
An optional feature provided in this module is the possibility to filter out classes in real-time upon request. In this way, the robot can map different objects online depending on the requirements proposed. Other implementation aspects will be further explained in Sect.~\ref{sect:experiments}.

\subsubsection{3D object position estimation using camera-lidar filtering}
This module fuses RGBD camera and lidar measurements to have a precise estimate of the objects' positions in the environment. The input is composed of the classification labels and masks found in the previous module, and depth information extracted from the camera (see Fig.~\ref{fig:pipeline}b) and the lidar (see Fig.~\ref{fig:pipeline}c). Sensors depth measurements are first analyzed separately in the following.

The depth image obtained from the camera (see Fig.~\ref{fig:pipeline}b) is filtered using the recognized objects masks through element-wise matrix multiplication. The output, containing only the depth data of the object plus some sensor noise and environment outliers, is used to build a 3D point cloud projecting the 2D image points in the 3D space using the formula in the equation:
\begin{equation} \label{eq:projection3d}
    \begin{bmatrix}
        x_C \\ y_C \\ z_C 
    \end{bmatrix}
    =
    \begin{bmatrix}
        \frac{1}{f_x} & 0  & -\frac{p_x}{f_x} \\
        0  & \frac{1}{f_y} & -\frac{p_y}{f_y} \\
        0  & 0  & 1
    \end{bmatrix}
    \begin{bmatrix}
        u \\ v \\ 1
    \end{bmatrix}
    z_C\text{ ,}
\end{equation}
where $x_C$, $y_C$, $z_C$ are the 3D point coordinates with respect to the camera, $u$, $v$ are the pixels on the image plane and $f_x$, $f_y$, $p_x$ and $p_y$ are the camera intrinsic parameters (focal distances and sensor's centre). Note that $z_C$ is the depth measured by the camera depth sensor.

The obtained point cloud is filtered using a voxel grid downsampling filter\footnote{voxel grid downsampling filter:~\url{https://pointclouds.org/documentation/tutorials/voxel_grid.html}} to reduce the number of points and, consequently, a radius outlier filter\footnote{radius outlier removal:~\url{https://pointclouds.org/documentation/tutorials/remove_outliers.html}} is applied to remove the outliers induced by sensors noises and inference imperfections. The final point cloud is then used to compute the camera artifact centroid $X_C$ as the mean of its points.

The 3D lidar centroid estimation is computed as follows. Projecting the 3D lidar points (see Fig.~\ref{fig:pipeline}c) in the 2D detected masks images using Eq.~\ref{eq:lidar_proj}, we are able to extract the object points of interest from the point cloud (\textit{i.e., } the points which have the 2D projection inside the mask). 
\begin{equation} \label{eq:lidar_proj}
    z_L
    \begin{bmatrix}
        u \\ v \\ 1
    \end{bmatrix}
    =
    \begin{bmatrix}
        f_x & 0  & p_x \\
        0  & f_y & p_y \\
        0  & 0  & 1
    \end{bmatrix}
    \begin{bmatrix}
        R_{L}^{C} & T_{L}^{C}
    \end{bmatrix}
    \begin{bmatrix}
        x_L \\ y_L \\ z_L \\ 1
    \end{bmatrix}\text{ ,}
\end{equation}
where $R_{L}^{C}\in\mathbb{R}_{3x3}$ and $T_{L}^{C}\in\mathbb{R}_{3x1}$ are the rotation matrix and the translation vector between the lidar and the camera, $x_L$, $y_L$, $z_L$ are the 3D centroid position with respect to the lidar and the other parameters are the same of Eq.~\ref{eq:projection3d}.

The extracted point cloud, representing the noisy artifact, will be then filtered using a radius outlier filter similar to the one used for the camera. Both radius filter parameters are directly dependent on the number of point cloud points because different distances and sizes of objects affect the point-cloud density and consequently the filtering. Finally, the mean of the point cloud is computed to obtain the lidar artifact centroid $X_L$.

\begin{figure}
    \centering
    \includegraphics[width=\linewidth]{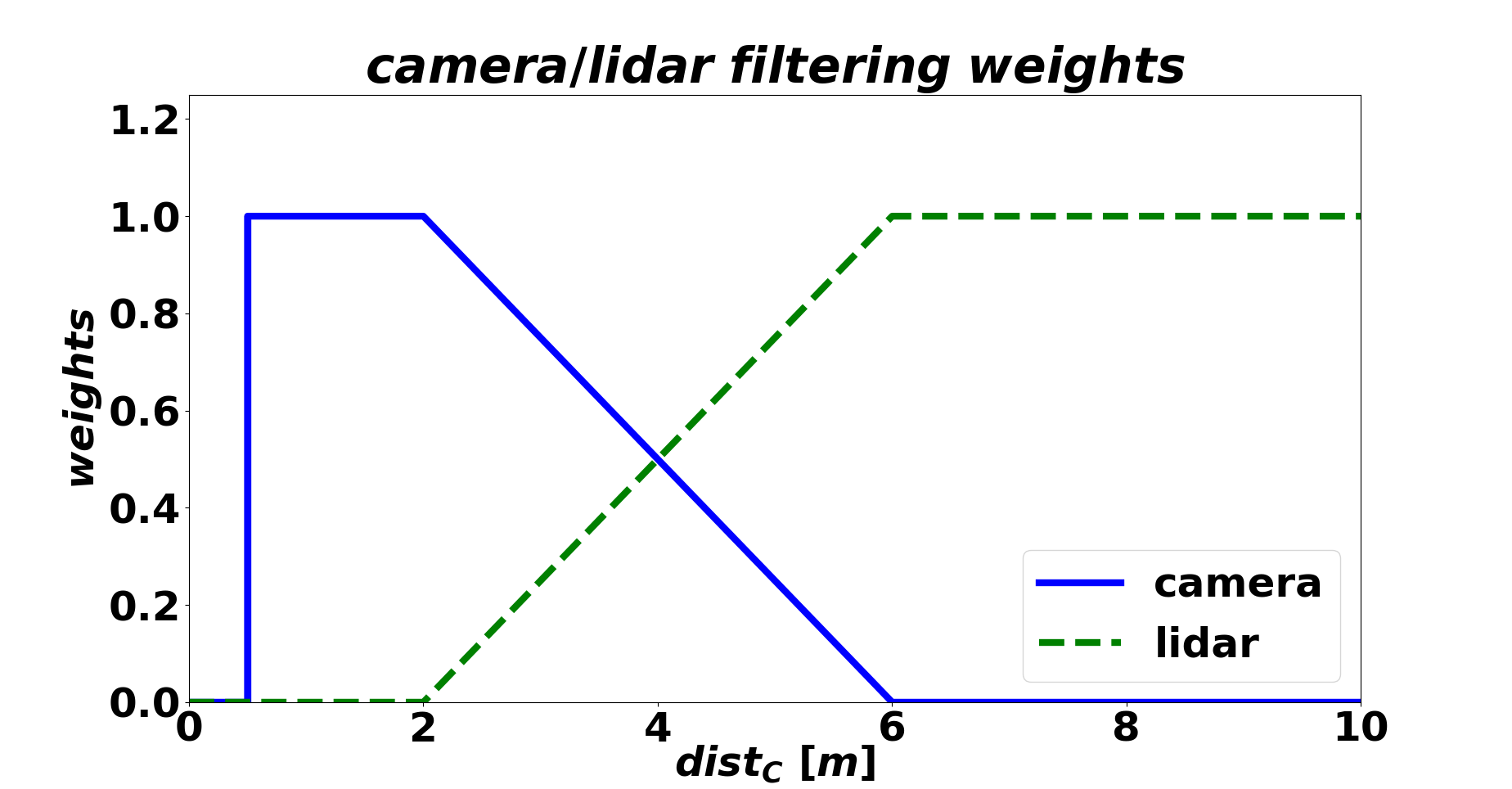}
    \caption{An example of the contribution weights of camera and lidar for sensor fusion. The camera weight is in blue while the lidar one is in dashed green. In the specific example, we considered the specifications of a generic RGB-D camera you can find on the market: $min_{C}=0.5$, $acc_{C}=2.0$ and $max_{C}=6.0$.}
    \label{fig:weight}
\end{figure}

Once both centroid measurements are available, they are fused in the artifact centroid $X$ following the rules in the equation:
\begin{equation} \label{eq:filtering}
    X=
    \begin{cases}
        \begin{tabular}{ll}
            $0$ & \text{If $ dist_{C} < min_{C} $ } \\
            $X_{C}$ & \text{If $ min_{C} \leq dist_{C} \leq acc_{C} $ } \\
            $\xi X_{C} + (1 - \xi) X_{L}$ & \text{If $ acc_{C} \leq dist_{C} \leq max_{C} $ } \\
            $X_{L}$ & \text{If $dist_{C} > max_{C}$}\text{ ,}
        \end{tabular}
    \end{cases}
\end{equation}
where $dist_{C}$ is the euclidean distance between the 3D point estimates and the camera, $min_{C}$ and $max_{C}$ are the minimum and maximum distances the depth camera can perceive, $acc_{C}$ is the distance within which the camera can have accurate enough measurements to be used alone for the object localization (the camera information are generally provided by the sensors vendors), $X_{L}\in \mathbb{R}^3$ and $X_{C}\in\mathbb{R}^3$ are the lidar and camera 3D centroid estimates and $\xi \in [0, 1] \in \mathbb{R}$ is the fusion weight represented by the blue slope of the segments between $acc_C$ and $max_C$ in Fig.~\ref{fig:weight} and it is computed as follows:
\begin{equation} \label{eq:xi}
    \xi = -\frac{1}{max_{C} - acc_{C}}(dist_{C}-acc_{C})+1
\end{equation}

Using the filtered camera and lidar point clouds, a rough 3D radius estimation $\rho$ of the objects is performed. The camera radius $\rho_{C}$ and the lidar radius $\rho_L$ are computed as the mean of the two bigger dimensions along the X, Y and Z point cloud axis. the final radius $\rho$ is computed following the same centroid fusion rules of Eq.~\ref{eq:filtering} substituting $X$ with $\rho$, $X_C$ with $\rho_C$ and $X_L$ with $\rho_L$.

Also, the view angle $\phi$ of the artifact with respect to the robot is computed. Such an angle is rotated with respect to the map reference frame for implementation reasons with equation:
\begin{equation} \label{eq:artifact_angle}
    \phi = atan2(r_{21}, r_{11}) + atan2(y_{r}, x_{r})\text{ ,}
\end{equation}
where the $r_{ij}$ is the entry at row $i$ and column $j$ of the rotation matrix $R_{r}^{m}\in\mathbb{R}_{3x3}$ between the map $m$ and the robot $r$ and $x_{r}$, $y_{r}$ are the x, y positions of the artifact centroid with respect to the robot base. The two addends of Eq.~\ref{eq:artifact_angle} represent respectively the heading angle between the robot and the map and the angle between the robot and the 3D centroid. 

\subsection{Artifacts manager for data association} \label{subsect:manager}

The manager (see Fig.~\ref{fig:pipeline}f) is needed to filter out outliers and to stabilize artifact position estimations provided by the sensor fusion module. This process is generally known as data association\cite{achour2022collaborative}\cite{xiang2017rnn}. The manager is composed of two modules: \textit{(i)} object position filtering and \textit{(ii)} object position stabilization which runs asynchronously in parallel.

\begin{center}
    \begin{table*} [hbt!]
        \centering
        \caption{Detection results of the simulation and real experiments.}
        \begin{tabular}{|c|ccc||ccc|}
            \hline
                & \multicolumn{3}{c||}{Simulation}  & \multicolumn{3}{c|}{Real} \\
            \hline
                & \textit{Camera} & \textit{Lidar}    & \textit{Fusion} & \textit{Camera} & \textit{Lidar} & \textit{Fusion} \\
            \hline
            \textbf{Correct detection}     & 386 & 391 & \bf{416 }    & 86   & 81   & \bf{99 } \\
            \textbf{Wrong localization}    & 12  & 13  & \bf{7 }       & 10   & 14   &  \bf{2 }  \\
            \textbf{Duplication}           & 19  & 24  & \bf{15 }       & 10   & 14   & \bf{6 }   \\
            \textbf{Wrong classification}  & \bf{0 }  & \bf{0}   & \bf{0  }     & 7    & 11   & \bf{6 } \\
            \hline\hline
            \textbf{Total detections}      & 417 & 428 & 433     & 113 & 120 & 113 \\
            \hline\hline
            \textbf{Total objects} & \multicolumn{3}{c||}{422}  & \multicolumn{3}{c|}{101}\\
            \hline
        \end{tabular}
        \label{tab:metrics}
    \end{table*}
\end{center}

\subsubsection{Position filtering}
Using a temporary data structure, the \textit{temporary buffer}, we store and filter the perceived artifacts. Once the manager receives the 3D artifacts position estimations from the perception module (see Sect.~\ref{subsect:perception}), it checks if the artifacts were already seen before (\textit{i.e.,} the distance between one of the already seen artifacts and the current one is less than its 3D radius). If this is the case then the artifact in the temporary buffer is updated. Otherwise, for each not previously seen artifact received, the manager creates a new artifact instance in the temporary buffer. These instances have their own moving average filter which estimates the average of the artifact centroid position and its radius with Eq.~\ref{eq:movingavg} and computes a variance based on the distances between the position and the moving average in the filter horizon with Eq.~\ref{eq:variance}.
\begin{align}
    &\mu = \frac{1}{N} \sum\limits_{\chi\in\Omega_N} \chi\label{eq:movingavg}\\ 
    &\sigma = \frac{1}{N} \sum\limits_{\chi\in\Omega_N} ||\chi - \mu||^2 \label{eq:variance}\text{ ,}
\end{align}
where $N\in\mathbb{N}$ is the number of measurement in the moving average set $\Omega_N$ of 3D points, $\chi\in\mathbb{R}^3$ represent the current 3D position measurement, $\mu\in\mathbb{R}^3$ is the 3D mean position and $\sigma\in\mathbb{R}$ represent the variance of the filter. 

\subsubsection{Position stabilization}
This module checks the stability of the artifacts in the temporary buffer and stores stable artifacts in another similar structure, the \textit{stable buffer}. If an artifact in the temporary buffer is stable, the stabilizer moves the artifact from the temporary buffer to the stable one. An artifact is considered stable when its moving average filter variance $\sigma$ is less than half its 3D artifact radius $\rho$ and at least half the average filter set $\Omega_N$ is filled. This means that we have enough stable object position estimations and the object position average can be used for fixing the object position on the map.

At the end of the Artifacts Mapping application, an additional data association step is performed. The artifacts belonging to the same class which overlay each other on the $XY$ plane are merged into a single artifact. This step reduces the duplicated object which sometimes appears on the map due to different point-of-view measurements and occlusions. After that, the stable artifacts buffer is saved in a yaml file which could be loaded into the user interface application presented in the next section. 

\subsection{User Interface for goal sending} \label{subsect:application}
A User Interface (UI) application based on a Rviz plugin (see Fig.~\ref{fig:artifact_example}) was developed to provide an intuitive visualization of the artifacts on the map, to send commands to the robot for moving near an artifact of interest and to delete artifacts which the user do not need or are wrong. Such artifacts can be loaded from the yaml file obtained with the artifacts mapping application. Through the UI application, the user can send \textit{nav\_msgs/goal} ROS messages which can be used by the robot to move towards the object (\textit{e.g.}, using the ROS navigation stack as we do, see Sect.~\ref{sect:experiments}). The user can interact with the artifacts by simply right-clicking on them on Rviz and selecting the action \textit{Go To} or \textit{Delete}. Being the artifacts centroid position inside the artifacts shapes, the goal is moved in front of the artifact so that the robot stops before colliding with the object. 
The other available option is artifact deletion. If the user notices that an artifact is wrongly identified (classification or position) then the user can delete it and, once the UI application is closed, the loaded yaml file is updated with the remaining artifacts.

\section{Experiments} \label{sect:experiments}

The experiments are performed both in simulation and using a real robot in a laboratory environment. The experimental setup is the same: some chosen objects are randomly positioned in the experiment area and the robot, following a predefined path, maps the predefined objects it encounters. This strategy is chosen because the objective is the validation of the artifacts mapping accuracy during an application, for example during a patrol. In other application scenarios, \textit{e.g.} search and rescue, our framework could run in parallel with an exploration algorithm and the robot could trigger the exploration module every time an object of interest is encountered to obtain a precise localization.

In the experiments, we compare the data fusion with the mono-sensors application (i.e. using only an RGB-D camera or only the lidar) to demonstrate that the data fusion highly improves the detection accuracy and decreases the errors. For each environment setup, the experiments are repeated three times, one for each \textit{sensors configuration}: only camera, only lidar, and both.

This work focuses only on semantic mapping and does not account for the robot localization which is assumed to be given. Additional errors in mapping resulting from localization are not considered in the final evaluation even if they negatively affect our application. Moreover, is important to notice that quadrupedal robots' movements are jerky and the sensors can suffer from that. 

We set the parameters $min_{C}$, $acc_{C}$ and $max_{C}$ of Eq.~\ref{eq:filtering} as $0.3$, $4$, $6$ respectively based on the camera hardware information provided by the camera vendors (Intel Realsense). 

The final validation performance is based on the number of objects which the robot can correctly find over the number of total objects. Also, the number of correctly-detected objects over the total number of detections is evaluated. 
The object is considered \textit{found} if the difference between the estimated position and the real one is less than the real object radius and the associated class label is correct. The errors are categorized as duplicated objects, wrong localization and wrong classification. The duplications occur when there are more artifacts on a single object. they could be caused by the wrong artifacts radius computation due to occlusions or distinct point of view detection (\textit{i.e., } viewed from different perspectives: front and behind). The localization is considered wrong if the artifact's estimated position is outside the real object shape while the classification is erroneous if the artifact's class label is not correct. 

For the simulation, the Whole-body Locomotion Framework (WoLF)\cite{raiola2022wolf} is used on a notebook with an \textit{Intel® Core™ i9-11950H} processor and an \textit{NVIDIA Geforce RTX 3080 Laptop} GPU. In the real scenario, a Unitree Go1\footnote{Unitree Go1:~\url{https://www.unitree.com/en/go1/}} quadrupedal robot equipped with a RoboSense RS-Helios16 lidar\footnote{RoboSense RS-Helios16:~\url{https://www.robosense.ai/en/rslidar/RS-Helios}}, an Intel RealSense D455\footnote{Intel RealSense D455:~\url{https://www.intelrealsense.com/depth-camera-d455/}} and three Nvidia Jetson\footnote{Nvidia Jetson:~\url{https://www.nvidia.com/it-it/autonomous-machines/embedded-systems/}} (two Jetson Nano 4GB and one Nvidia Xavier NX) are used for the evaluation.
The experiments are performed with the instance segmentation algorithms Yolact++~\cite{bolya2020yolact++} and YolactEdge~\cite{liu2021yolactedge} trained on COCO~\cite{lin2014microsoft} data set.

\begin{figure}
    \centering
    \includegraphics[width=\linewidth]{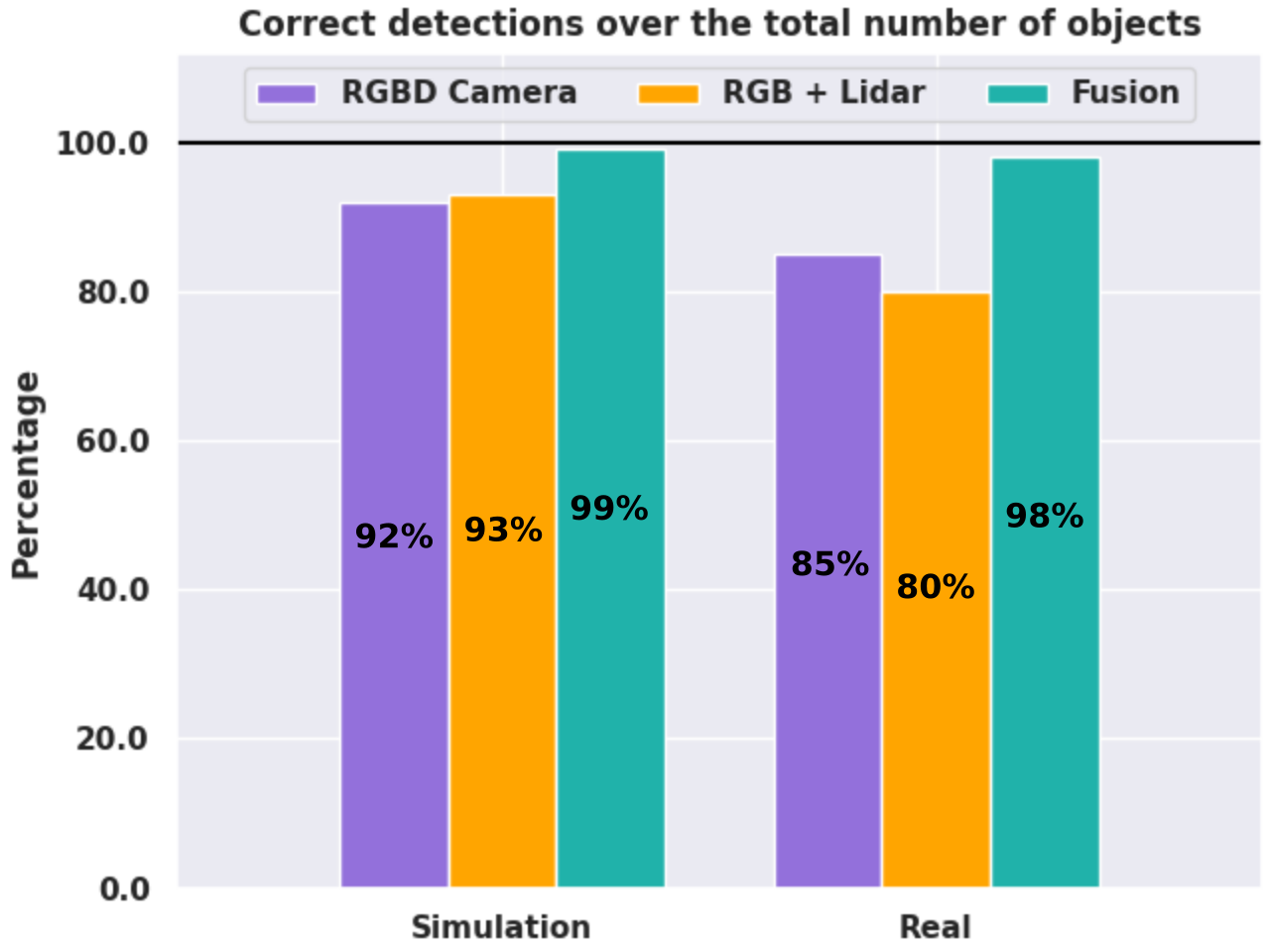}
    \caption{Percentage of the correctly mapped and labelled objects concerning the total number of objects on the scene. On the left are the simulation results and, on the right, are the real experiments. Each block has three histograms representing the three \textit{sensors configurations} used during the experiments: only RGBD camera, only RGB + lidar, and both.}
    \label{fig:correct_detections}
\end{figure}

\subsection{Simulation Experiments} \label{subsec:simulation}
Gazebo\footnote{Gazebo simulator: \url{https://gazebosim.org/home}} simulator is used to simulate the robot in two different environments: the office\footnote{Clearpath robotics worlds: \url{https://github.com/clearpathrobotics/cpr_gazebo/tree/noetic-devel/cpr_office_gazebo}} and Maze worlds where a predefined number of objects are positioned randomly at each iteration. The chosen objects for the simulation evaluation are \textit{vase}, \textit{couch}, \textit{plant} and \textit{person}. Specifically, in the office world, there are 5 vases, 12 couches, 6 plants and 11 persons while in the Maze world, there are 15 vases, 13 couches, 12 plants and 12 persons. The robot path is chosen randomly in advance using some waypoints on the map. In total, for each \textit{sensors configuration}, 10 experiments were conducted, 5 for each environment, using different setups, for a total of 30 experiments. 

\begin{figure}
    \centering
    \includegraphics[width=\linewidth]{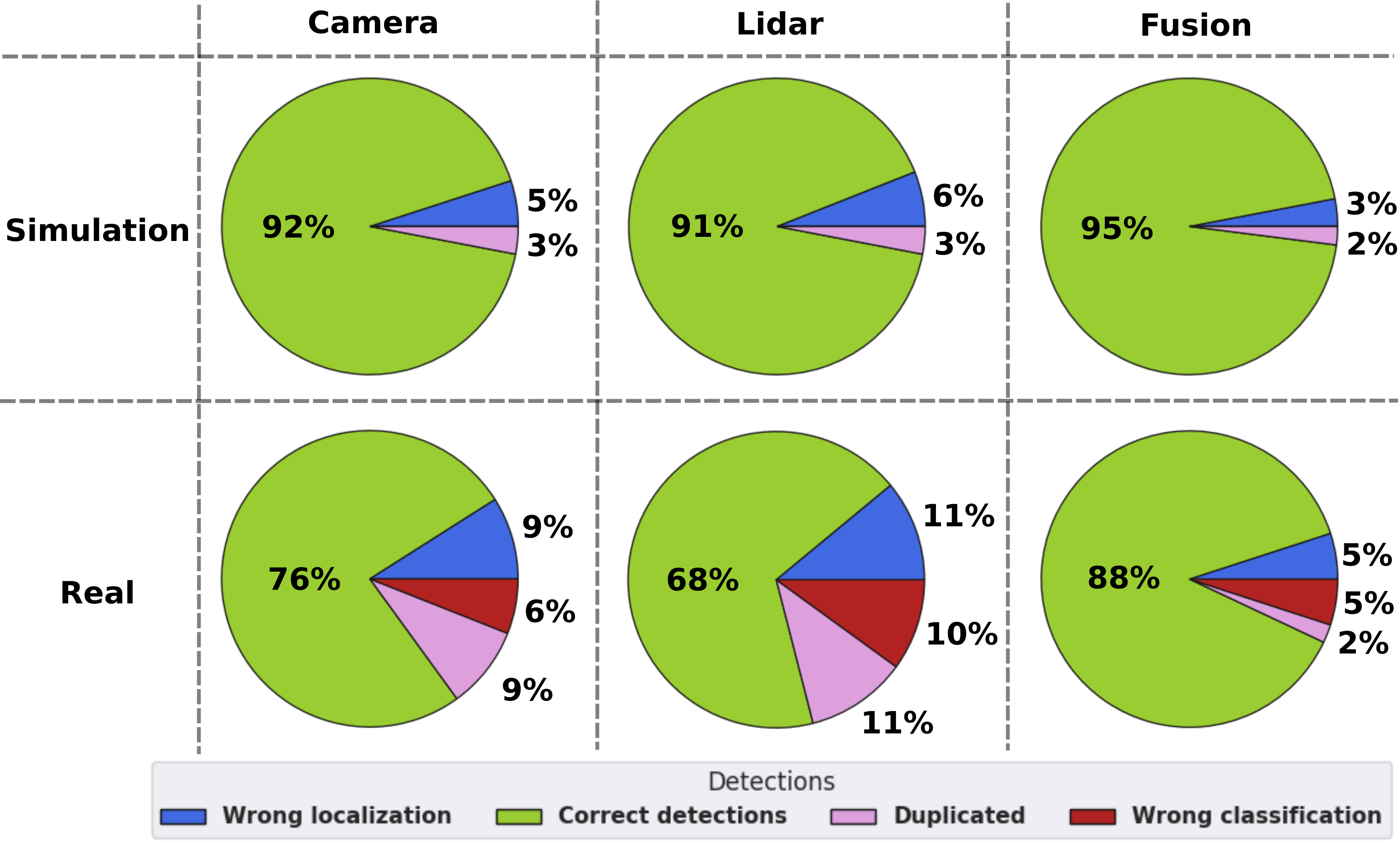}
    \caption{Distribution of correctly and wrongly detected artifacts among the total generated detections. The pie charts represent the distribution of the correctly detected artifacts in green, the doubled objects in blue, the wrongly localized ones in pink and the wrongly classified ones in red. On the top row are the simulations results while on the bottom are the real ones. For each row, experiments are divided into three columns depending on the \textit{sensors configuration} used during experiments: only camera, only lidar, and both.}
    \label{fig:total_detections}
\end{figure}

The results of the simulation experiments are shown in the left part of Fig.~\ref{fig:correct_detections} in terms of the number of correct detected objects. Specifically, considering the three ordered \textit{sensors configurations} (\textit{i.e.} only camera, only lidar, and both), we obtain the $92\%$, $93\%$ and $99\%$ of correctly localized and classified objects. Moreover, analysing the total number of detections produced, we obtain the distribution of the detections represented in the left column of Table~\ref{tab:metrics} and the top part of Fig.~\ref{fig:total_detections} for the simulation experiment. Among all the detection produced, considering again in order the three \textit{sensors configurations}, the $92\%$, $91\%$ and $95\%$ were correct while the remaining $8\%$, $9\%$ and $5\%$ of them were wrong. 

The farthest object correctly detected in simulation during the camera-lidar sensor fusion experiments was at $15.47m$ from the robot, while the nearest was at $1.23m$. 

\subsection{Laboratory Experiments} \label{subec:realexp}

The real experiments were carried out in a laboratory setting considering two scenarios, a one-room laboratory environment and a complete floor environment where the robot can move through corridors. In these environments were positioned \textit{umbrellas}, \textit{chairs}, \textit{cabinets}, \textit{backpacks} and \textit{TVs} in variable amounts. For each \textit{sensors configuration}, A total of 6 experiments were conducted, 3 for each environment, for a total of 18 experiments. For each trial, the objects were randomly moved and the illumination changed, \textit{i.e.}, switching off lights or closing shutters.

The results of the laboratory experiments are shown in the right part of Fig.~\ref{fig:correct_detections} in terms of the number of correct detected objects. Specifically, considering the three \textit{sensors configurations} in order (\textit{i.e.} only RGBD camera, only RGB + lidar, and both), we obtain respectively the $85\%$, $80\%$ and $98\%$ of correctly localized and classified objects. Moreover, analysing the total number of detections produced, we obtain the distribution of the detections represented in the right column of Table~\ref{tab:metrics} and the bottom part of Fig.~\ref{fig:total_detections} for the real experiment. Among all the detection produced, the $76\%$, $68\%$ and $88\%$ were correct while the remaining $24\%$, $32\%$ and $12\%$ of them were wrong. 

The farthest object correctly detected during the camera-lidar sensor fusion experiments was at a distance of $10.37m$ from the robot, while the nearest was at $0.98m$. 

\begin{figure}[t]
    \centering
\includegraphics[width=\linewidth]{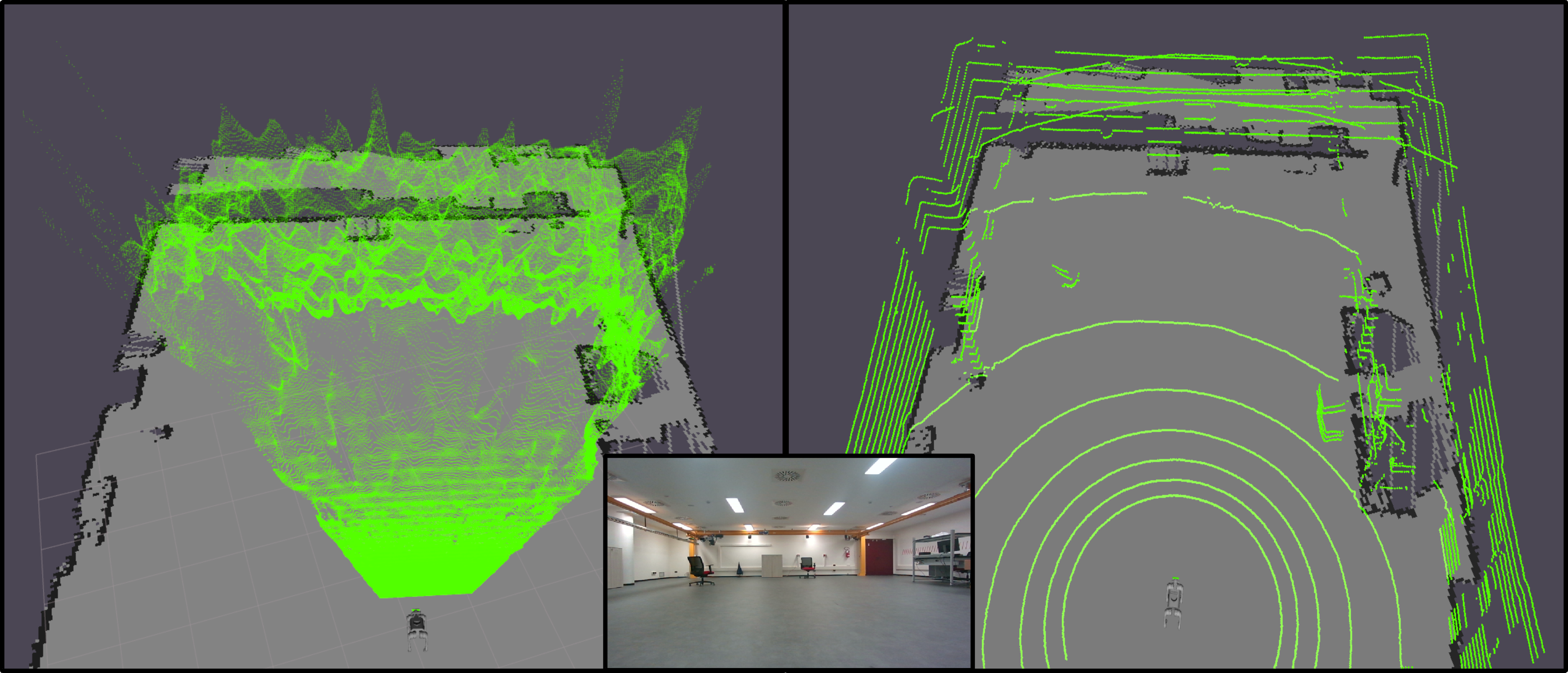}
    \caption{Qualitative comparison between RGB-D camera (left image) and lidar (right image) point cloud detections at an approximate distance of $10m$ from the wall. At the bottom centre, there is the representation of the scene taken with the robot camera at that time instant. At large distances, the camera data are noisier and less accurate with respect to the lidar one but at small distances cameras provide a denser accurate point cloud while lidar data are sparser. From this comparison can be deduced that a visual-lidar sensor fusion can enhance semantic mapping.}
    \label{fig:comparison}
\end{figure}

\subsection{Discussion} \label{subec:expdiscussion}

The first thing to point out is that the farthest distances of the detected object were greater than $10m$ both in simulation and in real experiments. We take into account this distance to show a qualitative comparison between the lidar and RGB-D measurement in Fig.~\ref{fig:comparison}. The figure qualitatively upholds the thesis that a lidar sensor along with the camera is necessary to improve semantic mapping and, in general, other detection algorithms in wide areas. 
Moreover, from the results obtained from the experiments, it is clear that in our framework the use of both sensors improves the robustness of the application and decreases the detection errors. These improvements are less evident in a simulation environment where we used almost ideal sensors, \textit{i.e.} the noise representation is not realistic as in Fig.~\ref{fig:comparison}. Still, it impacts real scenarios where there is more sensor noise. 

The lidar can map far obstacles precisely while the camera introduces lots of errors at high distances. If we adopt only the camera, one solution to avoid erroneous measurements could be to not consider the depth measurement out of the accurate range guaranteed by the device specifications. By the way, by doing this the robot could miss some artifacts if it does not get close enough to them. 

The camera, by providing more information at near distances with respect to the lidar, yields more precise centroid computations because it has fewer outliers than the lidar. Lidar outliers can be caused by wrong camera-lidar pose calibration and time synchronization which are essential for these applications especially when the robot moves fast. Instead, with RGBD cameras, the depth and the RGB images are synchronized in time and can be spatially superimposed almost exactly. 

It is important to notice that wrong classification errors result from erroneous classifications in the pre-trained instance segmentation neural network which can be caused by illumination, reflections or other environmental conditions. They are here considered because the image inference is a module of the proposed pipeline but such errors can be decreased using more powerful neural networks.

\section{Conclusion} \label{sect:conclusion}
We presented a framework which uses multi-modal sensors fusion to tackle the semantic mapping problem which is a rare setup in robotics applications. We fuse the lidar and RGB-D camera sensor readings to achieve better accuracy both for near and far objects as opposed to camera-only systems which lose accuracy for distant objects or lidar-only which lack high-level texture understanding of the environment. 

We proposed a UI application to interact with the artifacts map obtained during the mapping application. This application is useful to perform autonomous high-level decision-making tasks because it exposes the object's class and location to the robot and the user.

The experiments showed that our application can correctly detect, localize and map the $98\%$ of the objects present in the scene at different distances providing a small number of detection errors and good localization accuracy. The comparisons with the single-sensor scenario (only camera or only lidar) proved that sensor fusion is essential for wide areas and high-accuracy applications. 

There are different future improvements we planned for this framework: (i) evolve the algorithm to an independent graph-based SLAM system, (ii) use 3D semantic point clouds with oriented bounding boxes and dimension information for better visualization and object understanding, (iii) deal with dynamics obstacle.


{\small
\bibliographystyle{IEEEtran}
\bibliography{biblio}
}

\end{document}